\documentclass[lettersize,journal]{IEEEtran}
\usepackage{amsmath,amsfonts}
\usepackage{algorithmic}
\usepackage{algorithm}
\usepackage{array}
\usepackage{textcomp}
\usepackage{stfloats}
\usepackage{url}
\usepackage{verbatim}
\usepackage{graphicx}
\usepackage{cite}
\usepackage{booktabs}
\usepackage{multirow}
\usepackage{color}
\usepackage{subcaption}
\usepackage{hyperref}
\hypersetup{hidelinks}

\begin{document}

\title{BrainZ-BP: A Non-invasive Cuff-less Blood Pressure Estimation Approach Leveraging Brain Bio-impedance and Electrocardiogram}
\author{Bufang Yang, Le Liu, and Wenxuan Wu, Mengliang Zhou, Hongxing Liu, Xinbao Ning
}



\maketitle

\begin{abstract}
Accurate and continuous blood pressure (BP) monitoring is essential to the early prevention of cardiovascular diseases. 
Non-invasive and cuff-less BP estimation algorithm has gained much attention in recent years. 
Previous studies have demonstrated that brain bio-impedance (BIOZ) is a promising technique for non-invasive intracranial pressure (ICP) monitoring.
Clinically, treatment for patients with traumatic brain injuries (TBI) requires monitoring the ICP and BP of patients simultaneously.
Estimating BP by brain BIOZ directly can reduce the number of sensors attached to the patients, thus improving their comfort.
To address the issues, in this study, we explore the feasibility of leveraging brain BIOZ for BP estimation and propose a novel cuff-less BP estimation approach called BrainZ-BP.
Two electrodes are placed on the forehead and occipital bone of the head in the anterior-posterior direction for brain BIOZ measurement.
Various features including pulse transit time and morphological features of brain BIOZ are extracted and fed into four regression models for BP estimation.
Results show that the mean absolute error, root mean square error, and correlation coefficient of random forest regression model are 2.17 mmHg, 3.91 mmHg, and 0.90 for systolic pressure estimation, and are 1.71 mmHg, 3.02 mmHg, and 0.89 for diastolic pressure estimation. 
The presented BrainZ-BP can be applied in the brain BIOZ-based ICP monitoring scenario to monitor BP simultaneously.
\end{abstract}


\section{Introduction}
\IEEEPARstart{C}{ardiovascular} 
diseases (CVDs) have displaced communicable diseases as the major cause of global mortality \cite{10665-79059}.
Accurate and continuous
blood pressure (BP) monitoring is an effective tool to provide early prevention and management of CVDs.
Traditional BP measurement usually adopts sphygmomanometer or inflatable cuff-based oscillometer methods, which require applying pressure on the body of subjects by an inflatable cuff \cite{9528415}.
These cuff-based techniques are uncomfortable for users and are not suitable for monitoring circadian fluctuations of BP.
Therefore, continuous, non-invasive, and cuff-less BP monitoring techniques have sparked the research community's attention \cite{6725634,9142317,9406011}.

Numerous studies demonstrated that pulse wave velocity (PWV) is a prominent indicator to evaluate BP \cite{6256704}. 
PWV can be estimated by pulse transit time (PTT), which is the transit time of pulse wave propagation from two different skin sites. 
Several studies have investigated leveraging PTT for BP estimation, and their experimental results have proved that PTT is useful for BP estimation \cite{2020Investigation,6557001,7118672}. 
PTT computing usually requires at least two physiological signals, such as the electrocardiogram (ECG), photoplethysmogram (PPG), and bio-impedance (BIOZ).
Leveraging PPG and ECG signals is the most prevalent way for PTT computing \cite{7273864,7582472,7491263}.
However, PPG sensors consist of a light source and a photodetector, which is not only affected by environmental light and skin pigmentation but also consumes more power.
BIOZ is another non-invasive technique that can be used to compute PTT \cite{8857433}.
Pulsatile change of blood flow in each cardiac cycle causes the variation of BIOZ, thus BIOZ can be used for blood flow and respiration monitoring.
BIOZ avoids the shortcoming that PPG is limited by ambient light and power, and has gained a lot of attention in the cuff-less BP monitoring field \cite{8863984,8438854,s18072095}. 
Most existing cuff-less BP estimation approaches leveraging BIOZ signals are based on the wrist BIOZ \cite{8863984,8438854,s18072095,9247300,ibrahim2022cuffless}, ring BIOZ \cite{sel2023continuous} and chest BIOZ \cite{9547749}. 
There is currently a scarcity of studies exploring leveraging brain BIOZ for cuff-less BP estimation.

Brain BIOZ is also called rheoencephalography (REG) \cite{1995Rheoencephalographic}.
It is a valuable non-invasive technique for monitoring intracranial blood flow.
The most notable advantage of brain BIOZ lies in its non-invasive nature, allowing for its application in the non-invasive diagnosis of cerebral diseases, including intracranial hemorrhages \cite{8638835} and traumatic head injuries (TBI) \cite{e21060605}.
Brain BIOZ also plays a crucial role in intracranial pressure (ICP) monitoring \cite{PMID:26334594}. 
ICP measurement is an invasive measurement technique for the human brain, which requires using a probe to measure ICP in the context of craniotomy \cite{articleActa}.
As long-term invasive ICP measurement has the risk of intracranial infection, previous studies have begun to explore the utilization of brain BIOZ for non-invasive ICP monitoring \cite{Bodo_2015,2005BrainZ}.
In the intensive care units (ICU), patients with TBI need continuous monitoring of multiple physiological signals, not only ICP but also BP.
If BP can be directly estimated by brain BIOZ, the number of sensors attached to the patients can be reduced, thus improving patients' comfort.
However, existing research focuses on the utilization of PPG and wrist BIOZ for cuff-less BP estimation \cite{8863984,8438854,s18072095,9247300,ibrahim2022cuffless,sel2023continuous,9547749}, or investigates the correlation between brain BIOZ and ICP \cite{chen2022cerebral,Bodo_2015,2005BrainZ}.
Studies about the relationship between brain BIOZ and BP are missing, which is the motivation for this study.


To solve these limitations, in this study, we investigate the feasibility of using brain BIOZ for BP estimation for the first time and present a novel cuff-less BP estimation approach called BrainZ-BP. 
We summarize the contributions of this paper as follows:
\begin{itemize}		
	\item 
We implement a novel brain BIOZ-based BP estimation system. To the best of our knowledge, BrainZ-BP is the first work to investigate the feasibility of using brain BIOZ for non-invasive cuff-less BP estimation.
	\item 
	We develop a novel method for BP estimation based on brain BIOZ that encompasses considerations for signal processing, feature extraction, feature selection, and regression modeling.
	\item 
    We systematically investigate the effects of important parameters, including excitation frequency and electrode positioning on an individual's head, during brain BIOZ measurement on the accuracy of BP estimation.
	\item 
We implement BrainZ-BP and evaluate the performance through extensive experiments on a self-collected brain BIOZ dataset, and open-source the dataset at \href{https://github.com/bf-yang/BrainZ-BP}{https://github.com/bf-yang/BrainZ-BP}.
Results show that BrainZ-BP achieves a mean absolute error of 2.41 mmHg for systolic BP estimation and 2.01 mmHg for diastolic BP estimation, which can effectively applied in the brain BIOZ-based ICP monitoring scenario to monitor BP simultaneously.
\end{itemize}

\section{Background and Related work}

\subsection{Related Works}
\textbf{Brain BIOZ for Health Monitoring.}
Chen \emph{et al.} \cite{chen2022cerebral} investigated the correlation between brain BIOZ and cerebral blood flow (CBF), establishing a mathematical model that serves as a valuable tool for monitoring CBF based on brain BIOZ.
In addition, several previous studies have explored the utilization of brain BIOZ for non-invasive ICP monitoring.
Bodo \emph{et al.} \cite{Bodo_2015} investigated the correlation between brain BIOZ and ICP for rats.
They injected vinpocetine into rats to increase their CBF.
After injection, systemic arterial pressure of rats decreases about 25\% $\pm$ 14\%, and the amplitude of brain BIOZ and ICP signals both increases (BIOZ increase about 209\% $\pm$ 17\% and ICP increase about 28\% $\pm$ 16\%).
Traczewski \emph{et al.} \cite{2005BrainZ} studied the correlation between brain BIOZ and ICP obtained from the lumbar puncture.
They recruited 62 patients suspected of hydrocephalus in the experiments.
Results show that brain BIOZ has clinical value in the diagnosis and prognosis of hydrocephalus.

\textbf{BIOZ-based BP Estimation.}
Many studies have been proposed to utilize BIOZ for BP estimation.
Ibrahim \emph{et al.} \cite{8863984} leveraged wrist BIOZ and PPG for cuff-less BP estimation.
They employed the AdaBoost regression model for BP estimation based on the extracted PTT, time features, amplitude features, and area features.
Their approach achieves a root mean square error (RMSE) of 3.44 mmHg and a mean absolute error (MAE) of 2.51 mmHg for systolic blood pressure (SBP) estimation. 
For diastolic blood pressure (DBP) estimation, they achieved an RMSE of 2.63 mmHg and an MAE of 1.95 mmHg.
Huynh \emph{et al.} \cite{8438854} also leveraged wrist BIOZ and PPG signals for cuff-less BP estimation.
They established the relationship between PTT and BP by employing a quadratic regression model.
Their reported RMSE of SBP and DBP are 8.47 mmHg and 5.02 mmHg, respectively.
Huynh \emph{et al.} \cite{s18072095} proposed to only use BIOZ sensors for cuff-less BP estimation.
Two BIOZ sensors were placed on the participants' wrists to measure PTT. 
Subsequently, they utilized the PTT feature, along with the inverse relationship between PTT and PWV, to estimate blood pressure.
Wang et al. \cite{9247300} introduced a continuous BP monitoring system that leverages a single-channel wrist BIOZ signal. 
They developed a quadratic regression model for accurate BP estimation. 
The reported MAE for SBP and DBP were 2.01 mmHg and 2.26 mmHg, respectively.
Ibrahim et al. \cite{ibrahim2022cuffless} employed a wristband BIOZ sensor and a convolutional neural network (CNN) autoencoder for cuff-less BP estimation. 
Sel et al. \cite{sel2023continuous} developed a ring BIOZ device specifically designed for continuous cuff-less BP estimation, utilizing 15 BIOZ features and an AdaBoost regression model.

In summary, existing research focuses on the utilization of PPG and wrist BIOZ for cuff-less BP estimation, or investigates the correlation between brain BIOZ and ICP. 
However, studies exploring the feasibility of leveraging brain BIOZ for non-invasive cuff-less BP estimation are currently lacking.

\subsection{Application Scenario and Motivation}
Brain BIOZ is also called rheoencephalography (REG) \cite{1995Rheoencephalographic}. 
Blood flow in the brain changes periodically by the cardiac cycle, and the pulsatile change of blood flow causes the variation of BIOZ in the brain: electrical conductivity increases and impedance decreases when blood flows into the brain.
Brain BIOZ is a non-invasive technique that enables the monitoring of intracranial blood flow, rendering it valuable for diagnosing cerebral diseases such as intracranial hemorrhages \cite{8638835} and TBI \cite{e21060605}.
In addition, Brain BIOZ also plays a crucial role in ICP monitoring \cite{PMID:26334594}. 
Continuous and accurate monitoring of ICP is crucial for patients' health as long-term intracranial hypertension can lead to herniation, stroke, and even death. 
However, traditional ICP measurement is an invasive measurement technique for the human brain, where long-term invasive measurements can result in intracranial infection.

Brain BIOZ allows us to tackle this challenge from a new perspective.
Recent studies have shown that brain BIOZ is a promising technique for non-invasive ICP monitoring \cite{Bodo_2015,2005BrainZ}, as it can greatly reduce the risk of intracranial infection.
In the ICU, patients with TBI require continuous monitoring of multiple physiological signals, including ICP as well as BP. 
If BP can be directly estimated using brain BIOZ, it would allow for a reduction in the number of sensors attached to patients, thereby enhancing their comfort. 
However, existing studies \cite{8863984,8438854,s18072095,9247300,ibrahim2022cuffless} primarily focus on using wrist BIOZ for BP estimation, and there is a lack of research exploring the relationship between brain BIOZ and BP.  
This is also the motivation of this study.

\subsection{Principle of BIOZ Measurement}
\begin{figure}[!t]
	\centerline{\includegraphics[scale=0.32]{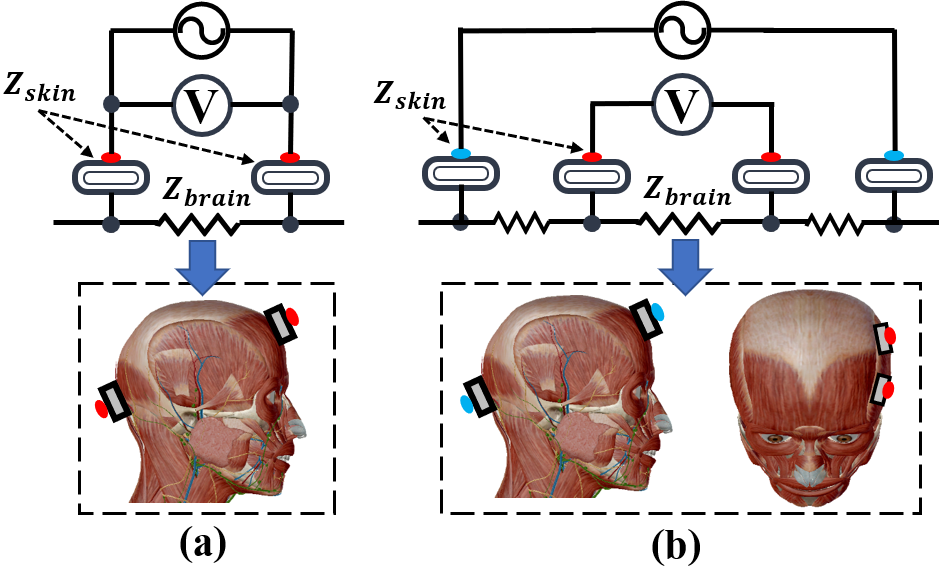}}
	\caption{Principle of four-electrode and two-electrode setups for brain BIOZ measurement. (a) Two-electrode setup, (b) four-electrode setup.}
	\label{fig_principle}
\end{figure}

According to the number of electrodes, BIOZ measurement is divided into two types, namely four-electrode and two-electrode setups \cite{8846577}. 
Fig. \ref{fig_principle} shows the brain BIOZ measurement principle of four-electrode and two-electrode setups.

Four-electrode setup uses two separate pairs of electrodes to inject high-frequency injection current and record voltage, respectively. 
It can obtain accurate impedance values as it avoids the effect of skin-electrode impedance ($Z_{skin}$) \cite{8320545}. 
In the four-electrode setup, the input impedance of the voltage measurement instrument is large enough, i.e. no current flows into the circuit of the voltage measurement instrument, so the effect of $Z_{skin}$ can be ignored. 
The rationale for four-electrode measurement is shown as follows:
\begin{equation}
V_S = I_c \times Z_{brain}
\label{eq3}
\end{equation}
where $I_c$ is the high-frequency injection current. 
$V_S$ is the voltage between two measuring electrodes. 
$Z_{brain}$ is the measured brain BIOZ. 
Four-electrode setup is suitable for scenarios that require accurate absolute values of BIOZ.
Ibrahim \emph{et al.} \cite{8863984} utilized four-electrode setup for wrist BIOZ measurement. 
Four electrodes were attached to the radial and ulnar arteries of the wrist.
Then they extracted PTT, time, and amplitude features from the wrist BIOZ signal, and used AdaBoost regression model for BP estimation. Wang \emph{et al.} \cite{9247300} presented a single-channel wrist-BIOZ-based system for BP estimation.
Four-electrode setup was used for BIOZ measurement in their study.
They used a current pump to provide a continuous excitation current with the frequency of 50 kHz and amplitude of 140 $\mu$A.

Two-electrode setup utilizes one pair of electrodes for current injection and voltage measurement simultaneously.
The rationale for two-electrode measurement is defined as follows:
\begin{equation}
V_S = I_c \times(Z_{brain} + 2 \times Z_{skin})
\label{eq4}
\end{equation}

Although the value of measured BIOZ is affected by $Z_{skin}$, two-electrode measurement can improve users' comfort and reduce the cost and complexity. 
Therefore, two-electrode technique is suitable for the scenario that requires the impedance variation rather than absolute impedance value \cite{9547749}.

\section{Materials and methods}

\subsection{System Architecture}
\begin{figure*}[!t]
	\centerline{\includegraphics[scale=0.33]{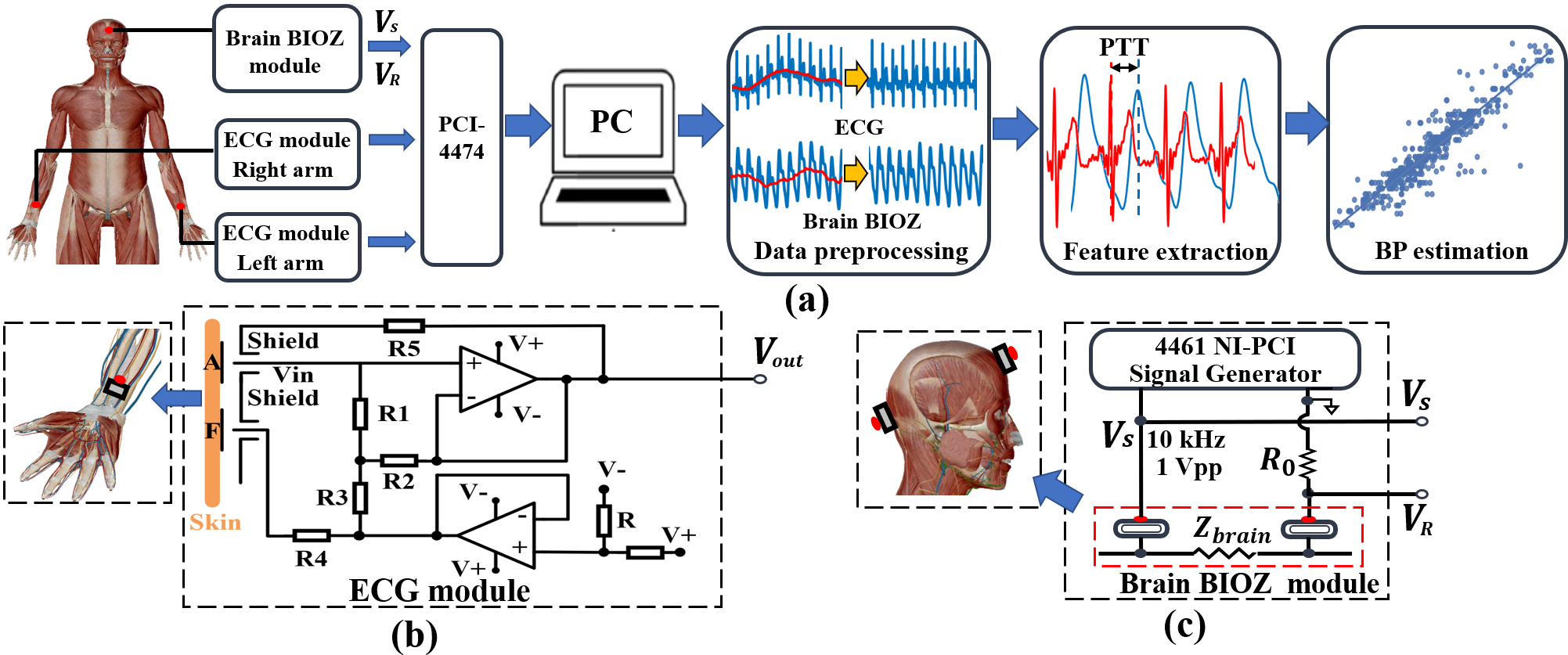}}
	\caption{Prototype of our proposed BrainZ-BP. (a) Overview of the BP estimation system. (b) Schematic of ECG measurement module. (c) Schematic of brain BIOZ measurement module.}
	\label{fig_system}
\end{figure*}

Our proposed BrainZ-BP contains a brain BIOZ measurement module and ECG measurement module, which are placed on the head, and the left and right wrists of subjects.
Brain BIOZ and ECG signals are recorded by PCI-4474 data acquisition (DAQ) card synchronously.
The measured $V_S$ and $V_R$ in the brain BIOZ measurement module are connected to the 
AI$_1$ and AI$_2$ of PCI-4474 DAQ card, respectively.
Excitation voltage $V_S$ is provided by signal generator PCI-4461.
The output of the ECG module is connected to the AI$_3$ of PCI-4474 DAQ card.
Several data preprocessing methods are utilized to remove the baseline wander and high-frequency noises in the measured brain BIOZ and ECG signals, where the sampling frequency is 100 kHz.
Then various features including PTT-based, and morphological features of brain BIOZ are extracted and fed into regression models for SBP and DBP estimation.
The prototype of the proposed BrainZ-BP is shown in Fig. \ref{fig_system}.

We use a modified two-electrode method for brain BIOZ measurement. 
A resistor $R_0$ is connected in series in the circuit.
We use a voltage resource $V_S$ and the resistor $R_0$ to replace the current source.
The rationale for brain BIOZ measurement can be written as follows:
\begin{equation}
V_S = V_R + \frac{V_R}{R_0} \times \left(Z_{brain} + 2 \times Z_{skin}\right)
\label{eq5}
\end{equation}
\begin{equation}
Z_{brain} \approx \left(\frac{V_S}{V_R}-1\right) \times R_0 = \left\{\frac{A_S}{A_R}e^{j(\phi_S-\phi_R)}-1\right\}\times R_0
\label{eq6}
\end{equation}
where $V_S$ is the high-frequency injected voltage. 
$V_R$ is the voltage of $R_0$.
$A_S$ and $A_R$ are the amplitudes of $V_S$ and $V_R$, and $\phi_S$ and $\phi_R$ are the phases of $V_S$ and $V_R$, respectively. 
In this study, $V_S$ is the sinusoidal voltage signal with an amplitude of 1 $V_{pp}$ and frequency of 10 kHz.
Numerous studies have shown the safety of applying a voltage signal with a frequency of 10kHz to the human head.
$R_0$ is the 10 k$\Omega$ resistor connected in series in the circuit.
The equivalent injected current flowing to the subject's head is 0.1 mA. 
Since the safe voltage and safe current of the human body are 24 V and 10 mA respectively, our circuit parameter settings are safe and reasonable.
Two electrodes are placed on the central line of the forehead and occipital bone of the human head in
the anterior-posterior direction for $Z_{brain}$ measurement.

ECG signal is recorded by a standard limb lead measurement method (Lead-I). 
Active electrodes are used in the ECG measurement module. 
The low output impedance of the active electrode reduces the effect of cable motion artifacts, and thus can improve the measurement performance of biosignal \cite{7828037}. 

\subsection{Data Pre-processing}
Signal qualities of brain BIOZ and ECG are threatened by four main factors: 
(1) baseline wandering due to the low-frequency respiration movement and variation of $Z_{skin}$, 
(2) power-line interference, 
(3) high-frequency excitation
voltage used for BIOZ measurement, and 
(4) high-frequency noises caused by body activities and muscular motion.
\subsubsection{Injection signal filtering}
Raw brain BIOZ and ECG contain 10 kHz excitation voltage. 
$N$ points segmental averaging method is utilized to eliminate this noise interference for ECG.
For brain BIOZ, we estimate the amplitude and phase of $V_S$ and $V_R$ for each $N$ points segment, then impedance can be computed by formula \eqref{eq6}. 
The new sampling rate is $1/N$ of the raw sampling rate. 
$N$ is set to 200, so the new sampling rates for ECG and brain BIOZ are both 500 Hz. 
Further, we utilize a 1000-order FIR bandpass filter (0.5-10 Hz) to eliminate power-line interference and high-frequency noises.

\subsubsection{Baseline calibration}
ECG and brain BIOZ contain baseline wandering. 
Savitzky-Golay (SG) filter is utilized to smooth and denoise the two signals in this study \cite{1964Smoothing}. 
The advantage of SG filter is that the length and the size of output data can remain the same as the input signal, while noises can be eliminated. 
In this study, order and window size are set to 3 and 10001 (about 20 seconds), respectively. 

\subsubsection{Segmentation}
In this study, a sliding window is utilized to perform data segmentation. 
Experimental results of previous studies have demonstrated that an 8 seconds window with 6 seconds overlapping is capable to extract key characteristics of cardiac activity \cite{9082808}. Therefore, 8 seconds sliding window with 75\% overlapping is used in this study to split the raw 30-second data into multiple data segments. 

\begin{figure}[!ht]
	\centerline{\includegraphics[scale=0.225]{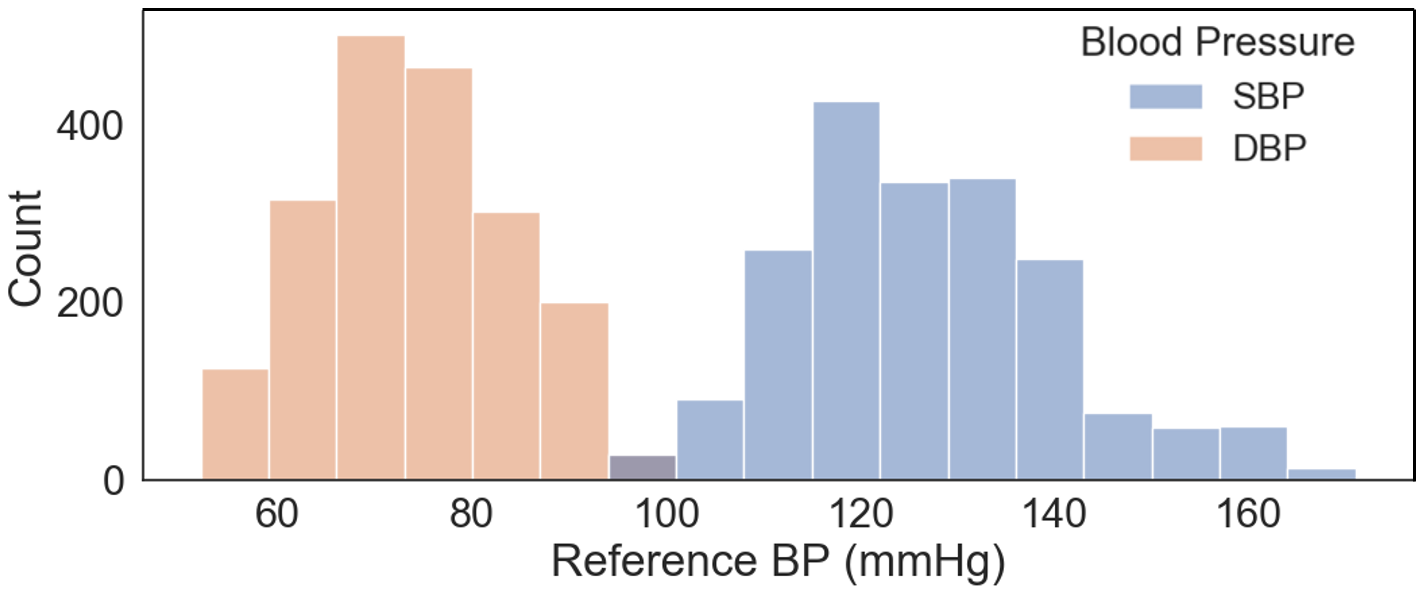}}
	\caption{Statistical distribution of SBP and DBP in the database.}
	\label{fig_distribution}
\end{figure}

Finally, the database contains 1942 ECG and brain BIOZ recordings acquired from 13 subjects. 
Each 8-seconds recording has two labels (reference SBP and DBP). 
The mean values of SBP and DBP in the database are 126.3 $\pm$ 14.6 mmHg and 73.3 $\pm$ 10.2 mmHg, respectively. 
Fig. \ref{fig_distribution} shows the statistical distribution of reference SBP and DBP in the database. 

\subsection{Feature Extraction}
\begin{figure}[!t]
	\centerline{\includegraphics[scale=0.33]{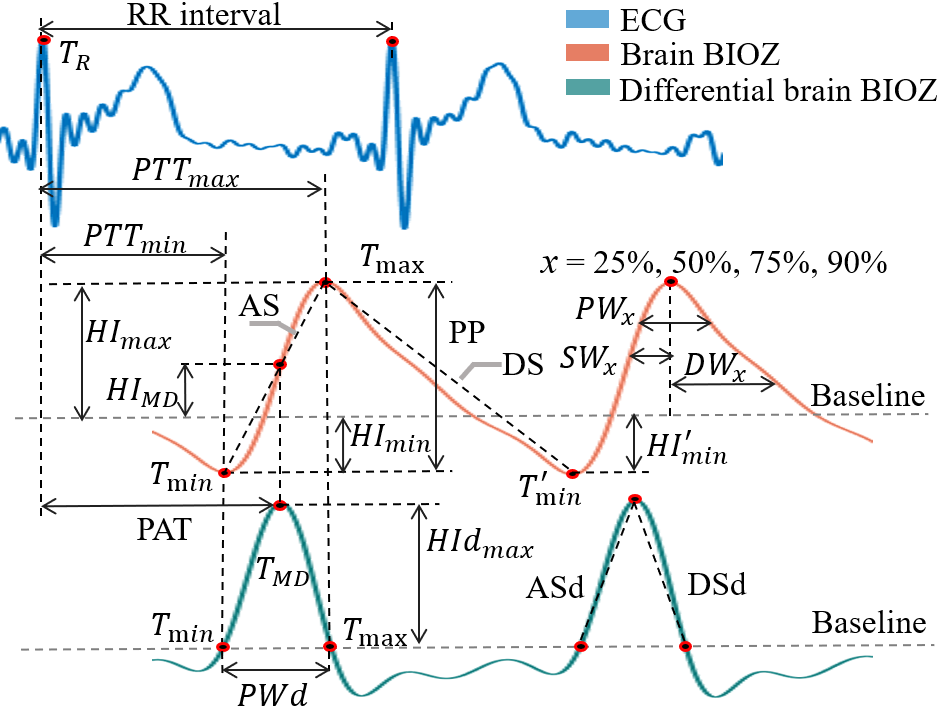}}
	\caption{Schematic diagram of PTT-based features, morphological features, height features, slope features and differential signal features.}
	\label{fig_feature_extraction}
\end{figure}

A total of 42 features, including cardiac cycle-based and segment-based features, are extracted from ECG and brain BIOZ in this study.
For segment-based features, we calculate them in each 8 seconds data segment.
For cardiac cycle-based features, we calculate them in each cardiac cycle and compute the mean as the corresponding features.
Fig. \ref{fig_feature_extraction} shows the schematic diagram of 
extracted features.
The definition of the 42 features used in this study is listed in Table \ref{table2}.
\begin{table}[t]
	\centering
	\caption{Definition of the extracted features in this study.}\label{table2}
	\resizebox{\columnwidth}{!}
	{
	\begin{tabular}{ccc}
		\toprule
		No.& Features & Description \\
		\midrule
		1 & PTT$_{max}$ & Time delay between R peak of ECG and peak of brain BIOZ \\
		2 & PTT$_{min}$ & Time delay between R peak of ECG and minimum point of brain BIOZ \\
		3 & PAT & 
		Time delay between R peak of ECG and MD point of brain BIOZ\\
		
		\midrule
		4 & DW & Diastolic width of brain BIOZ \\
		5 & DW$_{25}$ & \multirow{4}*{\shortstack{Diastolic width at x\% of the peak of brain BIOZ\\x = 25, 50, 75, 90, respectively}} \\
		6 & DW$_{50}$ &  \\
		7 & DW$_{75}$ &  \\
		8 & DW$_{90}$ &  \\	
		
		\midrule
		9 & SW & Systolic width of brain BIOZ \\
		10 & SW$_{25}$ & \multirow{4}*{\shortstack{Systolic width at x\% of the peak of brain BIOZ\\x = 25, 50, 75, 90, respectively}} \\
		11 & SW$_{50}$ &  \\
		12 & SW$_{75}$ &  \\
		13 & SW$_{90}$ &  \\
		
		\midrule
		14 & PW & Pulse width of brain BIOZ \\
		15 & PW$_{25}$ & \multirow{4}*{\shortstack{Pulse width at x\% of the peak of brain BIOZ\\x = 25, 50, 75, 90, respectively}} \\
		16 & PW$_{50}$ &  \\
		17 & PW$_{75}$ &  \\
		18 & PW$_{90}$ &  \\
		
		\midrule
		19 & PWR$_{25}$ & \multirow{4}*{Ratio of pulse width at x\% of the peak to total pulse width} \\
		20 & PWR$_{50}$ &  \\
		21 & PWR$_{75}$ &  \\
		22 & PWR$_{90}$ &  \\
		
		\midrule
		23 & HI$_{max}$ & Maximum height of brain BIOZ \\
		24 & HI$_{min}$ & Minimum height of brain BIOZ \\
		25 & HI$_{\rm MD}$ &MD point height of brain BIOZ \\
		26 & PP & Height difference between HI$_{max}$ and HI$_{min}$ \\
		27 & HIR$_{max}$ & Ratio of HI$_{max}$ and HI$_{min}$ \\
		28 & HIR$_{\rm MD}$ & Ratio of HI$_{\rm MD}$ and HI$_{min}$ \\		
		
		\midrule
		29 & AS & Ascending slope of brain BIOZ \\
		30 & DS & Descending slope of brain BIOZ \\
		
		\midrule
		31 & HId$_{max}$ & Maximum height of differential brain BIOZ \\
		32 & PWd & Pulse width of differential brain BIOZ \\
		33 & PWd$_{50}$ & Pulse width at 50\% of the peak of differential brain BIOZ \\
		34 & PWRd & Ratio of PWd$_{50}$ to PWd \\
		35 & ASd & Ascending slope of differential brain BIOZ \\
		36 & Dsd & Descending slope of differential brain BIOZ \\
		
		\midrule
		37 & SD & Standard deviation of brain BIOZ \\
		38 & Skew & Skewness of brain BIOZ \\
		39 & Kurt & Kurtosis of brain BIOZ \\
		
		\midrule
		40 & ApEn & Approximate entropy of brain BIOZ \\
		41 & SampEn & Sample entropy of brain BIOZ \\
		\midrule
		42 & HR & Heart rate \\
		\bottomrule
	\end{tabular}
	}
\end{table}

\subsubsection{PTT-based features}
PTT-based features are considered to be useful in BP estimation \cite{2020Investigation,6557001,7118672}. 
In this study, three important points in brain BIOZ are utilized to compute PTT features, namely, maximum point, minimum point, and maximum derivative (MD) point. 
Based on the three points and R peak of ECG, PTT$_{max}$, PTT$_{min}$, and PAT are calculated in each cardiac cycle, which are defined as follows:
\begin{equation}
PTT_{max} = T_{max} - T_R
\label{eq8}
\end{equation}
\begin{equation}
PTT_{min} = T_{min} - T_R
\label{eq9}
\end{equation}
\begin{equation}
PAT = T_{MD} - T_R
\label{eq10}
\end{equation}
where $T_{max}$, $T_{min}$, $T_{MD}$ and $T_R$ are the time of maximum point, minimum point, MD point of brain BIOZ, and of the R peak of ECG in the current cardiac cycle, respectively.

\subsubsection{Morphological features}
Morphological features of brain BIOZ can reflect the cardiovascular condition, which is crucial for BP estimation \cite{8863984}.
Morphological features consist of pulse width (PW), systolic width (SW), and diastolic width (DW) of brain BIOZ, which are defined as follows:
\begin{equation}
PW = T^{\prime}_{min} - T_{min}
\label{eq11}
\end{equation}
\begin{equation}
SW = T_{max} - T_{min}
\label{eq12}
\end{equation}
\begin{equation}
DW = T^{\prime}_{min} - T_{max}
\label{eq13}
\end{equation}
where $T^{\prime}_{min}$ is the minimum point of brain BIOZ in the next cardiac cycle.
PW, SW, and DW at 25\%, 50\%, 75\%, and 90\% of the peak of brain BIOZ are extracted in each cardiac cycle, which are denoted as PWx, SWx, and DWx, respectively. 
Further, the ratio of PWx and the total width PW are calculated, which is denoted as PWRx.

\subsubsection{Height features}
Height features of brain BIOZ, including maximum height (HI$_{max}$), minimum height (HI$_{min}$), MD point height (HI$_{\rm MD}$), and peak to peak (PP) value are extracted for each cardiac cycle. 
Further, height ratio features HIR$_{max}$ and HIR$_{\rm MD}$ are extracted, which are defined as:
\begin{equation}
HIR_{max} = \frac{HI_{max}}{HI_{min}}
\label{eq14}
\end{equation}
\begin{equation}
HIR_{\rm MD} = \frac{HI_{\rm MD}}{HI_{min}}
\label{eq15}
\end{equation}

\subsubsection{Slope features}
Slope features are considered to be useful for BP estimation since they can reflect the velocity of CBF. 
Two slope features are computed in this work, namely, ascending slope (AS) and descending slope (DS). 
AS and DS represent the slope in the systolic area and diastolic area, respectively, which are defined as follows:
\begin{equation}
AS = \frac{HI_{max}-HI_{min}}{T_{max}-T_{min}}
\label{eq16}
\end{equation}
\begin{equation}
DS = \frac{HI_{max}-HI^{\prime}_{min}}{T_{max}-T^{\prime}_{min}}
\label{eq17}
\end{equation}

\subsubsection{Statistical features}
In this study, three statistical features including standard deviation (SD), skewness (Skew), and kurtosis (Kurt) are extracted for each 8-s brain BIOZ segment. 

\subsubsection{Entropy features}
Entropy features are useful quantization indicators for complexity and irregularity of time series \cite{8370620}. 
Approximate entropy (ApEn) and sample entropy (SampEn) of brain BIOZ are extracted in this study. 

\subsubsection{Differential signal features}
First-order difference of brain BIOZ is an effective quantization indicator for the velocity of blood flow in the human brain. 
Maximum height (HId$_{max}$), pulse width (PWd), pulse width at 50\% height of the peak (PWd$_{50}$), and the ratio of PWd$_{50}$ and PWd are calculated in each cardiac cycle. 
We also extracted the ascending slope (ASd) and descending slope (DSd) features, which can reflect the acceleration of CBF. 

PTT-based features, morphological features, height features, slope features, and differential signal features are cardiac cycle-based features, while statistical features and entropy features are segment-based features.

\subsection{Feature Importance Analysis}
\label{feature_analysis}
Hand-crafted features may contain redundant information.
To gain a deeper understanding of the importance of hand-crafted features in BP estimation,  we consider two types of feature importance analysis techniques feature importance evaluation, namely Pearson correlation coefficient (PCC) and random forest impurity.
PCC and random forest impurity are linear and non-linear techniques for feature importance evaluation.
However, whether to employ either of these methods individually or combine them for feature selection, leading to a more accurate BP estimation, is still an open problem.
In this study, we compare three approaches for feature selection, namely PCC, random forest impurity, and their combination.

\subsubsection{Pearson correlation coefficient}
PCC is an effective technique for feature importance analysis, which can reflect the linear correlation between each feature and the corresponding predicted value \cite{6563175}. 
PCC is defined as follows:
\begin{equation}
r_{xy}=\frac{\sum_{i=1}^{N}\left\{ (x_i-\mu_x)\cdot (y_i-\mu_y) \right\}}{\sqrt{\sum_{i=1}^{N}(x_i-\mu_x)^2}\cdot \sqrt{\sum_{i=1}^{N}(y_i-\mu_y)^2}}
\label{eq24}
\end{equation}
where $x_i$ and $y_i$ are the extracted feature variable and the BP value of the $i$-th sample, respectively. 
$\mu_x$ and $\mu_y$ are the mean of the feature and the mean of the BP value in $N$ data samples.
$r_{xy}$ is between -1 and 1. 
A closer absolute value of $r_{xy}$ to 1 means a higher linear correlation.
According to the $abs(r_{xy})$ of each feature, PCC importance ranking is obtained (larger absolute $r_{xy}$ ranks higher).

\subsubsection{Random forest impurity}
We also utilize a non-linear feature importance evaluation method called random forest impurity \cite{9142317}. 
Random forest impurity method uses the impurity of each feature to evaluate the feature importance during tree model recursively building, which is an embedded feature selection method \cite{KHAIRE2019}.

Random forest (RF) is an ensemble tree model, which integrates multiple Classification and Regression Trees (CART) \cite{Qi2012}. 
The construction process of each CART is essentially a process of feature selection. The objective function of each CART is as follows:
\begin{equation}
Obj\!=\!\mathop{min}\limits_{j,s}\!\left\{
\!\mathop{min}\limits_{c_1}\!\sum_{x_i\in R_1}\!(y_i-c_1)^2\!+\!
\mathop{min}\limits_{c_2}\!\sum_{x_i\in R_2}\!(y_i-c_2)^2\! 
\right\}
\label{eq25}
\end{equation}
Based on the $j$-th feature and feature splitting point $s$, data is split into two partitions, namely $R_1$ and $R_2$. 
For each partition $R$, the predicted value is denoted as $c=\frac{1}{n}\sum_{i\in R}y_i$. 
Mean square error (MSE) is used as loss function to find the optimal splitting feature and splitting point. 
Suppose $N$ is the total number of 
samples, $M$ is the total times of $X$ used as the splitting feature in CART, the importance of feature $X$ is defined as:
\begin{equation}
I_C(X)\!=\!\sum_{i=1}^{M}\!\left\{\!
P\!\left(\!R^{(i)}\!\right)\!N_R^{(i)}\!-\!P\!\left(\!R_l^{(i)}\!\right)\!N_{R_l}^{(i)}\!-\!P\!\left(\!R_r^{(i)}\!\right)\!N_{R_r}^{(i)}\!
\right\}
\label{eq26}
\end{equation}
where $R^{(i)}$ is the partition before the $i$-th splitting using feature $X$.
$R_l^{(i)}$ and $R_r^{(i)}$ are the left and right partitions after the $i$-th splitting. 
$N_R^{(i)}$, $N_{R_l}^{(i)}$ and $N_{R_r}^{(i)}$ are the number of samples of partitions $R^{(i)}$, $R_l^{(i)}$ and $R_r^{(i)}$, respectively.
$P(\cdot)$ represents the impurity of the partition, i.e. MSE value.

Let $D$ be the number of CART in RF, the importance of feature $X$ in the random forest is defined as:
\begin{equation}
I_{RF}(X)=\frac{1}{D}\sum_{i}^{D}I_i(X)
\label{eq28}
\end{equation}
where $I_i(X)$ is the importance of feature $X$ in the $i$-th CART.

\subsubsection{Combining two methods for feature selection}
To evaluate the linear and non-linear correlation between each feature and BP value simultaneously, we first use PCC and RF impurity methods respectively to calculate their respective feature importance ranking.
Then, the average of two rankings is used as the feature importance score.
According to the importance score, the final ranking is obtained.

\subsection{Regression Models}
Numerous machine learning (ML) algorithms have been proposed and widely employed for health monitoring systems \cite{yang2021single,yang2022novel,yang2020automatic}.
In this work, four ML models are used for BP estimation, including linear regression (LR), support vector machine (SVM), decision tree (DT), and random forest (RF). 
Details about these models are as follows:
\subsubsection{Linear regression}
LR is a frequently used regression model, which builds the linear relationship between the input feature vector and the predicted variable. 
It is regarded as a baseline model in our experiment. 

\subsubsection{Support vector machine}
SVM is a powerful statistical machine learning algorithm, which is also called SVR when it is applied in regression task \cite{Cristianini2000AnIT}. 
The training strategy of SVR model is based on the structural risk minimization principle. 
SVR maps the data from low-dimension space into high dimension space and searches for an optimal hyperplane.
It is suitable for small-sample learning problems.

\subsubsection{Decision tree and Random forest}
The advantage of DT lies in its high interpretability and low computational cost \cite{8531733}. 
RF is an ensemble model, integrating multiple decision trees, which belongs to bagging ensemble learning. 
It utilizes bootstrap sampling, introducing sample variation and feature variation, thereby reducing the variance of the base model. 
Suppose there are $D$ CARTs in the RF model, and the output of RF can be expressed as follows:
\begin{equation}
RF(x)=\frac{1}{D}\sum_{i=1}^{D}CART_i(x)
\label{eq30}
\end{equation}
where $CART_i(x)$ is the predicted value of the $i$-th CART. 

\subsection{Experimental Protocol}
The experiment was conducted under the IRB approval (2022DZKY-040-01) by Nanjing Jinling Hospital.
A total of 13 healthy subjects without any history of cardiovascular diseases were recruited for our experiment (11 males, 2 females, age: 23.6 $\pm$ 1.6 years, height: 171.5 $\pm$ 6.7 cm, weight: 65.3 $\pm$ 12.6 kg). 
The mean body mass index (BMI) of these subjects are 22.0 $\pm$ 2.9 kg/$m^2$.

\begin{figure}[!t]
	\centerline{\includegraphics[scale=0.49]{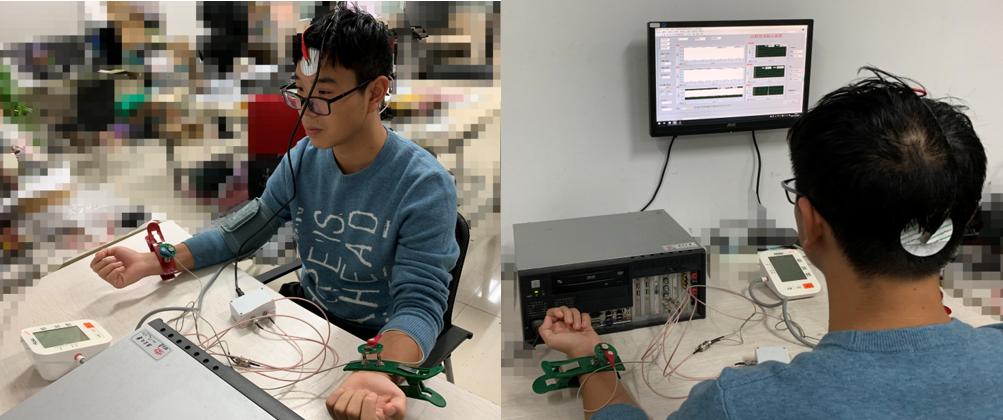}}
	\caption{Experimental scenario.}
	\label{fig_prototype}
\end{figure}

A total of 10 measurement trials are conducted for each subject. 
Each trial lasts for 30 seconds. 
During each trial, ECG, brain BIOZ, reference SBP, and DBP, are recorded synchronously for each subject. 
Fig. \ref{fig_prototype} shows the experimental scenario. 
During data measurement, participants are required to sit quietly and remain still to avoid signal noises and motion artifacts.
Reference SBP and DBP values are recorded by a cuff-based BP device (BSX-533, Haier, China), which is placed on the right upper arm. 

Since BP will not change largely if subjects just sit on the seat quietly. 
To obtain a wider range of BP values, subjects are required to conduct physical exercises in the experiment.
Before the fifth measurement trial started, subjects were required to conduct high knee exercises for 2 minutes. 
After high knee exercise, the BP value of each subject will increase largely, where SBP will increase about 10-30 mmHg, and DBP will increase about 5-10 mmHg. 
We also try to instruct subjects to conduct other exercises such as the squat, but it is harmful to their knees, and is difficult to increase DBP value via this exercise. 
After high knees exercise, subjects are required to remain still again during the sixth to tenth measurement trials.
The BP value of subjects will gradually decrease until it returns to its normal level.
In this way, the BP value of subjects will first increase and then decrease during the 10 trials, so a wider range of BP values are recorded. 

\subsection{Model Development and Performance Evaluation}
A total of 1942 recordings from 13 subjects were collected in our experiments. 
We perform 10-fold cross-validation to evaluate the BP estimation performance of our method. 
In this study, RF contains 500 decision trees, and the minimum number of samples in the leaf node is 1. 
We also experimented with the number of trees ranging from 10 to 1000. Results show that the estimation performance improves obviously when the tree number increases from 10 to 500, but the performance can hardly improve when the tree number is larger than 500. 
Therefore, the tree number is set to 500 in our experiment. Radial basis function (RBF) kernel is used for SVR model, and the regularization parameter is set to $10^3$ in our experiments.

Mean error (ME), mean absolute error (MAE), root mean square error (RMSE), and correlation coefficient (R) are used to evaluate the BP estimation performance in this study. 
ME and RMSE are utilized to assess whether the BP estimation model satisfies the Association for the Advancement of Medical (AAMI).
MAE is used to assess the BP estimation model in accordance with the British Hypertension Society (BHS) standard.

\section{Experiments and Results}
\subsection{Brain BIOZ Waveforms}
\begin{figure}[!t]
	\centerline{\includegraphics[scale=0.45]{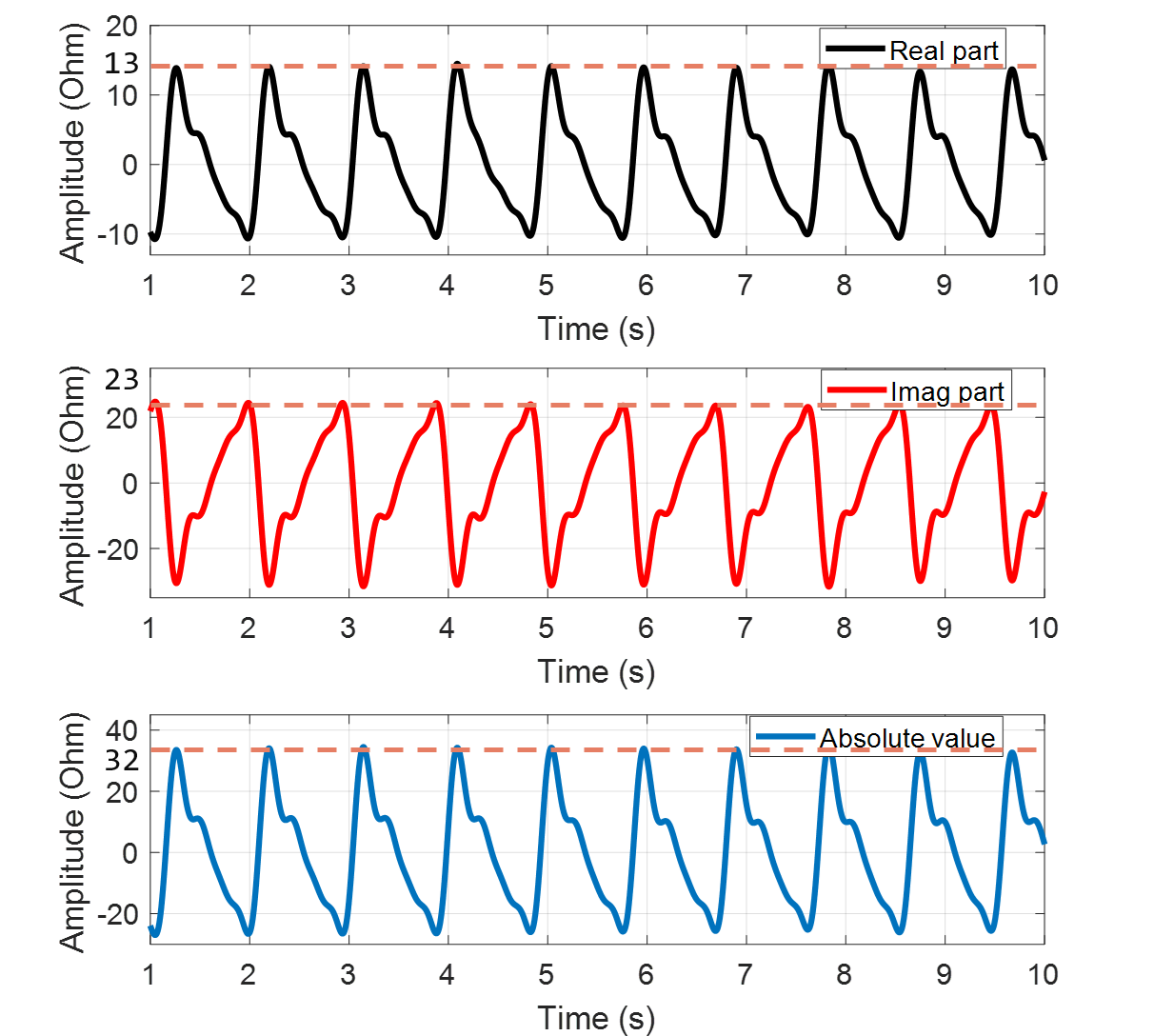}}
	\caption{
 The processed brain BIOZ signal.}
	\label{Real_Imag}
\end{figure}

Fig. \ref{Real_Imag} shows the example of measured brain BIOZ when two electrodes are placed on the forehead and occipital bone of the head in an anterior-posterior direction.
From top to bottom are the real part, imagine part, and the absolute value of the brain BIOZ signal. 
As can be seen, the absolute value of brain BIOZ has the highest amplitude (about 32 $\Omega$), followed by the image part and the real part (about 23 $\Omega$ and about 13 $\Omega$, respectively). 
The phenomenon that the absolute value of brain BIOZ provides the largest impedance change also occurs in other data segments.
A larger amplitude of brain BIOZ can better reflect CBF change.
Therefore, the absolute value of brain BIOZ is used for analysis in this study.

\subsection{Feature Importance Analysis}
\begin{figure}[!ht]
    \centering
    \begin{subfigure}{1\columnwidth}
        \centering
        \includegraphics[width=1\textwidth]{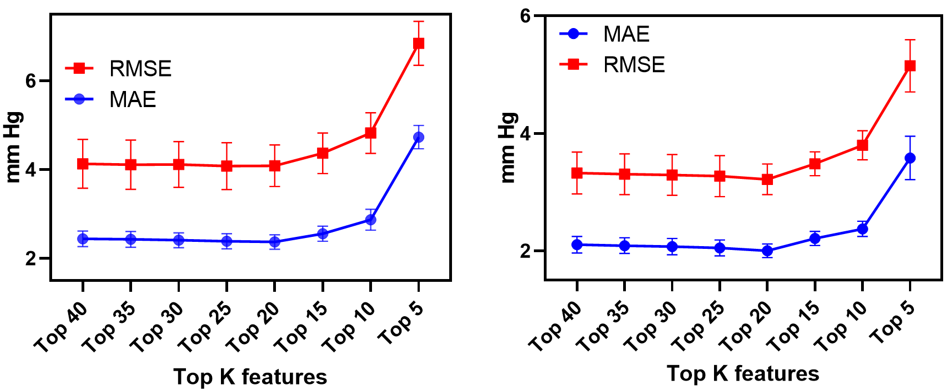}
	\caption{PCC. The left and right figures are SBP and DBP estimations, respectively.} \label{feature_selection_PCC}
    \end{subfigure}

    \vskip\baselineskip

    \begin{subfigure}{1\columnwidth}
        \centering
        \includegraphics[width=1\textwidth]{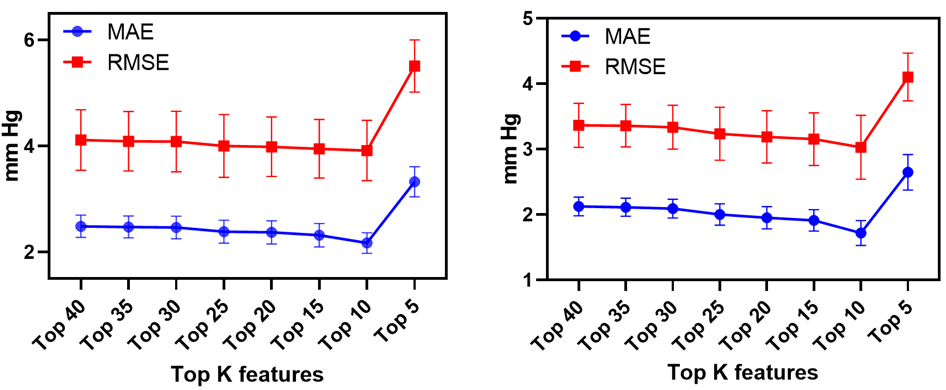}
	\caption{Random forest impurity. The left and right figures are SBP and DBP estimations, respectively.
 } \label{feature_selection_RF}
    \end{subfigure}

    \vskip\baselineskip

    \begin{subfigure}{1\columnwidth}
        \centering
        \includegraphics[width=1\textwidth]{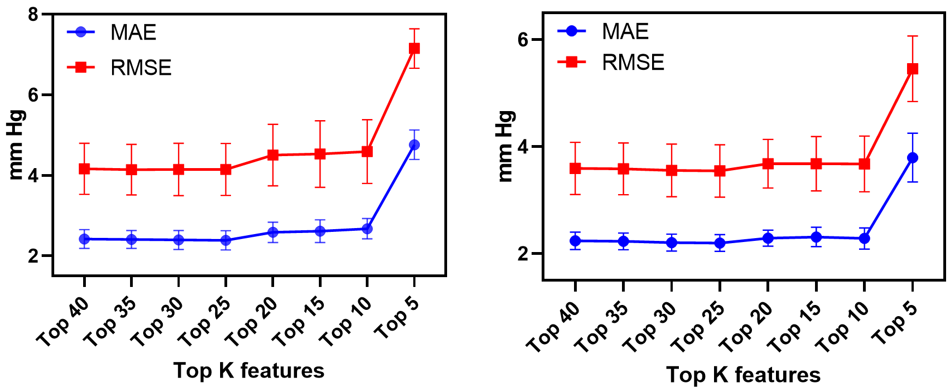}
	\caption{Combination of PCC and random forest impurity. The left and right figures are SBP and DBP estimations, respectively.} \label{feature_selection_combine}
    \end{subfigure}
    
  \caption{
  BP estimation performance using three types of methods to select the top K important features. (a) and (b) are from PCC. 
  (c) and (d) are from random forest impurity. 
  (e) and (f) are from the combination of PCC and random forest impurity.
  }
  \label{feature_select}
\end{figure}

Fig. \ref{feature_select} shows the BP estimation performance using three types of methods to select the top K important features.
For each K, a 10-fold cross-validation experiment is conducted.
It is observed that the estimated error of PCC and random forest impurity both exhibit a trend of initially decreasing and then increasing as the number of top K features used is reduced (Fig. \ref{feature_selection_PCC} and Fig. \ref{feature_selection_RF}).
 This can be attributed to the fact that irrelevant and redundant features can be harmful to machine learning algorithms \cite{shukla2019feature,liu2013iterative}, i.e. increasing the prediction error and training speed.
Furthermore, random forest impurity exhibits a more pronounced trend of decreasing and then increasing compared to PCC. It also achieves a lower estimation error for both SBP and DBP at its optimal point.
However, when employing the feature selection method that combines PCC and random forest impurity (as introduced in Section \ref{feature_analysis}), it demonstrates an almost monotonic increasing trend.
The optimal point is reached when utilizing the top 25 important features, but the estimated error is higher compared to using random forest impurity alone for feature selection.

For ease of comparison, we have summarized the best results from each approach in Table \ref{feature_select_table}.
We can see that using random forest impurity for feature selection achieves the lowest BP estimation error and requires the fewest number of features. While integrating PCC and random forest impurity provides a comprehensive perspective on feature importance (linear and non-linear relationships), combining the two approaches does not yield better results and even leads to worse performance. Therefore, we opt for random forest impurity as the feature importance ranking strategy in BrainZ-BP, as it demonstrates superior performance and efficiency in feature selection.
Based on the results in Fig. \ref{feature_selection_RF}, the top 10 important features selected by random forest impurity are used in BrainZ-BP.

\begin{table}
	\centering
	\caption{The best result of each approach and the number of features they used (K value).}\label{feature_select_table}
	\begin{tabular}{p{0.35cm}cccc}
		\toprule
		& Evaluation   & \multirow{2}*{PCC}    & Random forest  & Combi- \\
  		& Metrics   &    & impurity &nation \\
		\midrule		
		\multirow{2}*{SBP} 
		& RMSE (mmHg) & 4.08 $\pm$ 0.44 & \textbf{3.91 $\pm$ 0.53} & 4.14 $\pm$ 0.61 \\
		& MAE (mmHg) & 2.37 $\pm$ 0.15 & \textbf{2.17 $\pm$ 0.18} & 2.38 $\pm$ 0.22 \\
		\midrule
		\multirow{2}*{DBP} 
		& RMSE (mmHg) & 3.21 $\pm$ 0.24 & \textbf{3.02 $\pm$ 0.46} & 3.54 $\pm$ 0.46 \\
		& MAE (mmHg) & 2.00 $\pm$ 0.11 & \textbf{1.71 $\pm$ 0.18} & 2.19 $\pm$ 0.14 \\
  
            \midrule
            \multicolumn{2}{c}{Number of features} & 20 & \textbf{10} & 25\\
  
		\bottomrule
	\end{tabular}
\end{table}


\begin{figure}[!t]
	\centerline{\includegraphics[scale=0.31]{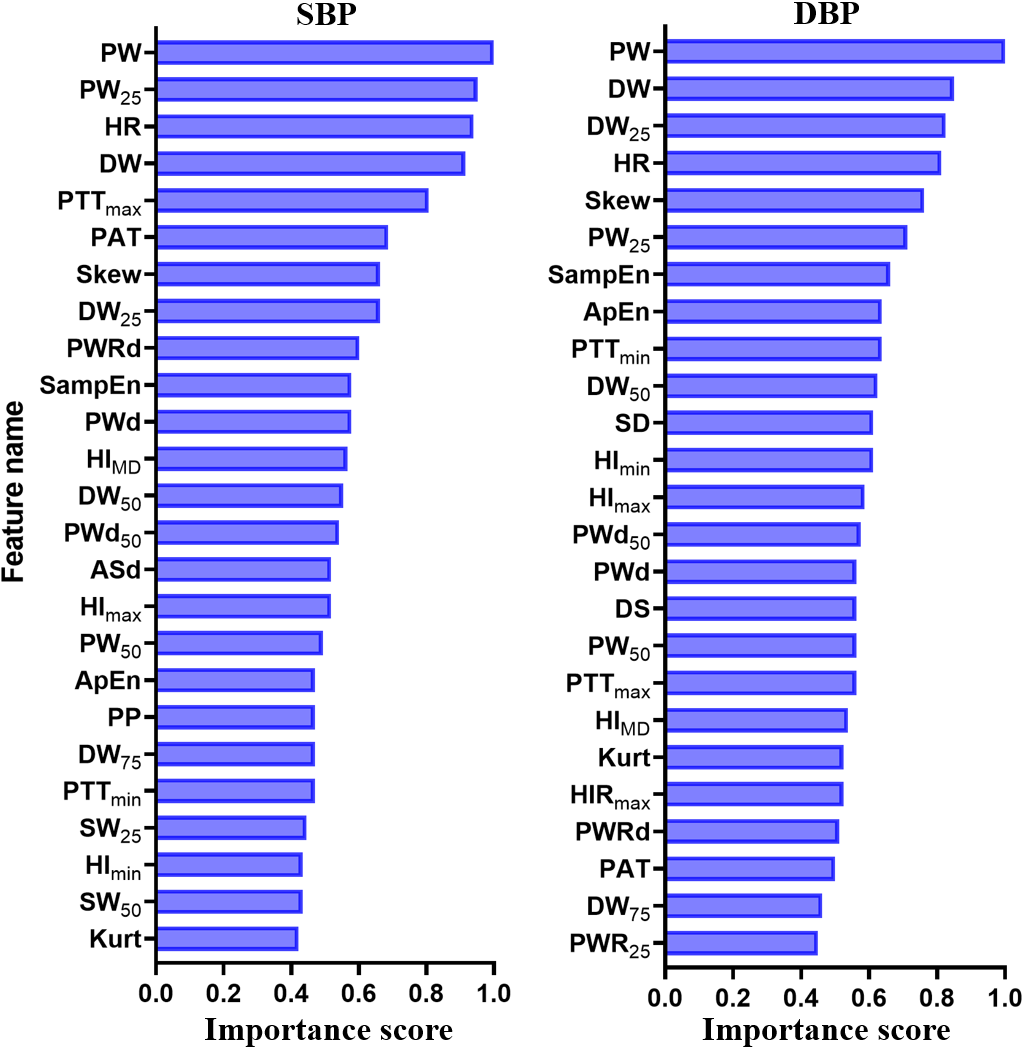}}
	\caption{Feature importance ranking for SBP and DBP estimation.}
	\label{feature_importance}
\end{figure}

Feature importance ranking for SBP and DBP estimation are shown in Fig. \ref{feature_importance} (only the top 25 most important features are listed).
As can be seen, morphological features of brain BIOZ, e.g. PW, PW$_{25}$, DW, DW$_{25}$, play a crucial role both in SBP and DBP estimation, which is consistent with study \cite{9528415}.
The pulse width of brain BIOZ (PW) has the highest importance score both for SBP and DBP estimation.
Additionally, PTT-based features including PTT$_{max}$, PTT$_{min}$, PAT, are also important for both SBP and DBP estimation, as previously shown in studies \cite{2020Investigation,6557001,7118672,7273864}.

\subsection{Performance of BP Estimation}
\begin{figure}[!t]
	\centerline{\includegraphics[scale=0.28]{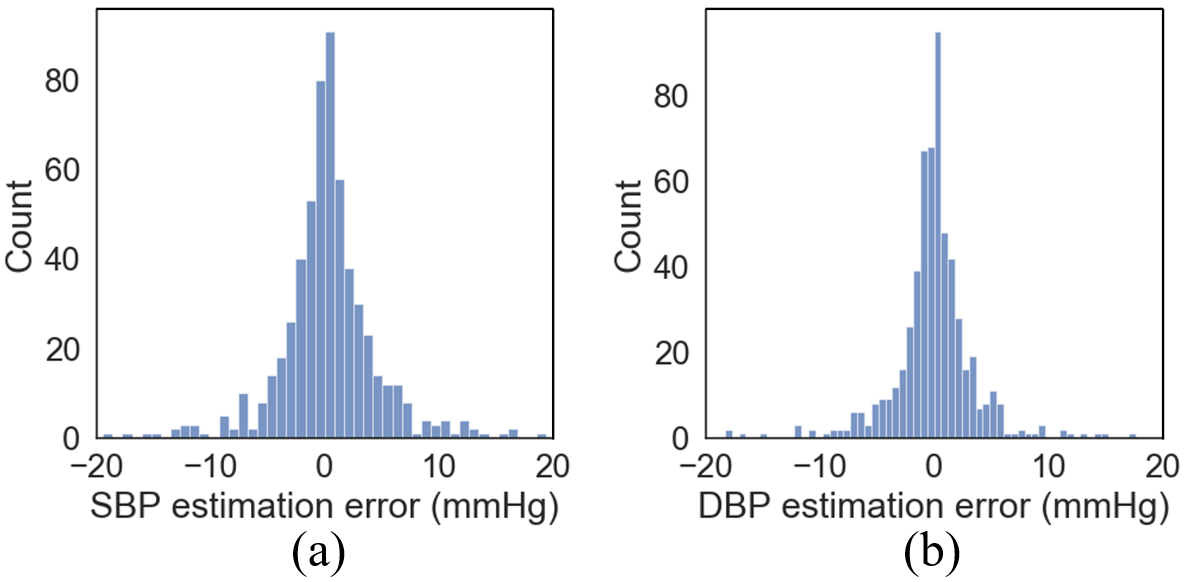}}
	\caption{Histograms of estimation error using RF model. (a) SBP estimation. (b) DBP estimation.}
	\label{Error_histogram}
\end{figure}

\begin{figure}[!t]
	\centerline{\includegraphics[scale=0.28]{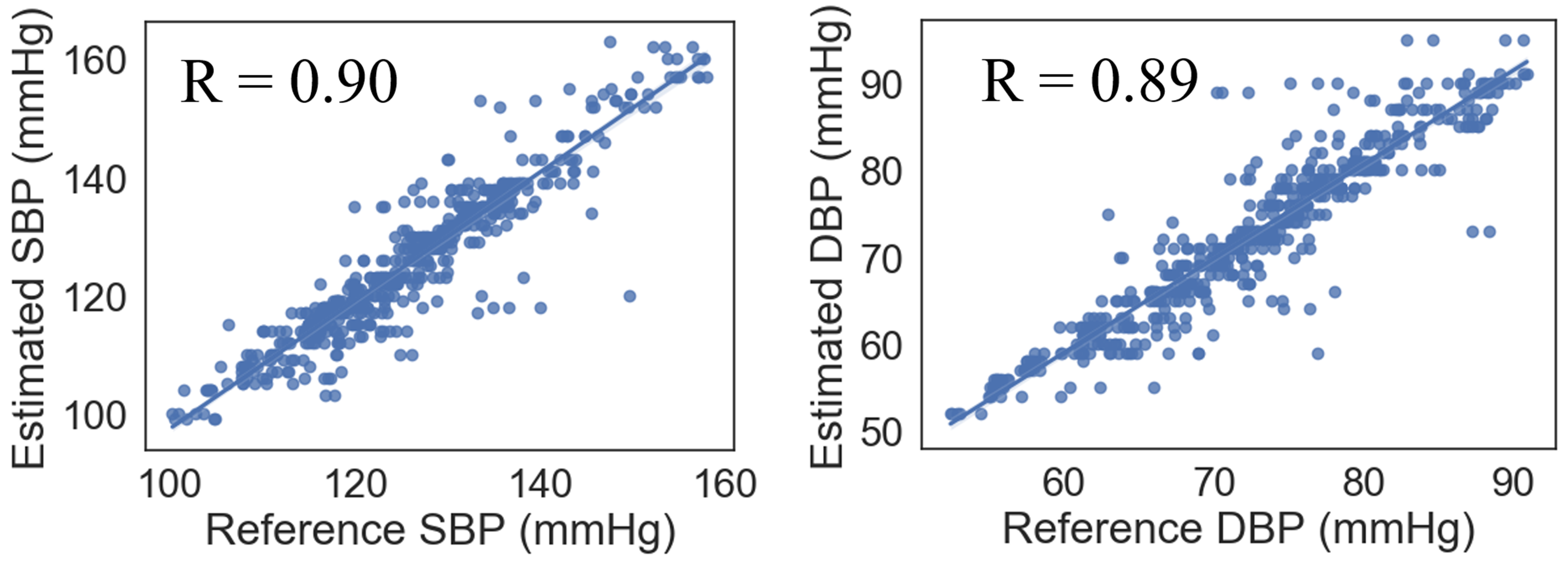}}
	\caption{Correlation plots of estimated BP with reference BP.}
	\label{Corr_R}
\end{figure}

\begin{figure}[!t]
	\centerline{\includegraphics[scale=0.28]{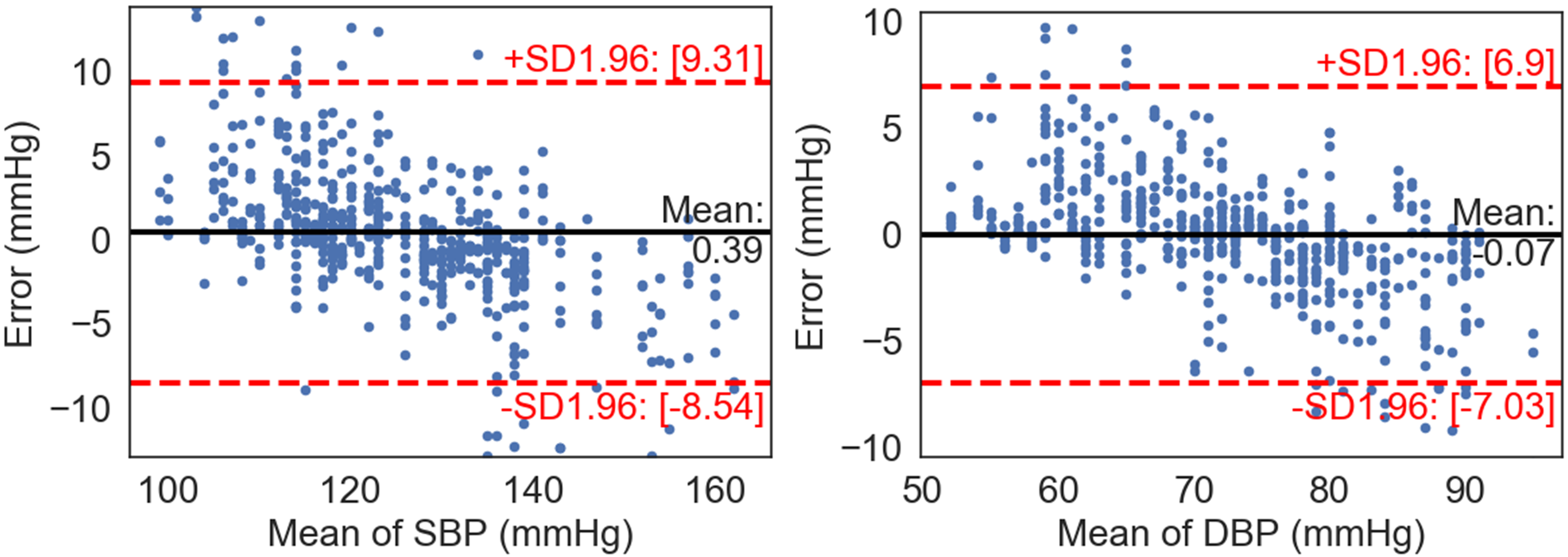}}
	\caption{Bland-Altman plots of SBP and DBP estimation.}
	\label{BlandPlot}
\end{figure}

Fig. \ref{Error_histogram} shows the histograms of SBP and DBP estimation errors using RF. 
As can be seen from the histograms, most of the predicted errors are distributed around zero, within the range of $\pm$ 20 mmHg, which is similar to the Gaussian distribution with zero means.
Fig. \ref{Corr_R} presents the correlation plots of estimated BP with reference BP using RF.
Correlation coefficient R is 0.90 and 0.89 for SBP and DBP estimation, which illustrates the estimated BP of our method is highly correlated with reference BP. 
Fig. \ref{BlandPlot} shows the Bland-Altman plots of SBP and DBP estimation. X-axis and Y-axis are the mean and error of reference BP and estimated BP respectively.
Bland-Altman plots illustrate that most of the predicted errors of SBP and DBP are within 0.39 $\pm$ 8.92 mmHg and -0.07 $\pm$ 6.97 mmHg limits. 
Therefore, our proposed approach is an effective BP estimation method.

Additionally, we compare the BP estimation performance of the four regression models, as shown in Table \ref{table5}. 
As can be seen, the estimation performance of LR model is obviously lower than the other three regression models. 
R is only 0.41 and 0.30 for SBP and DBP estimation, respectively. 
Amongst the other three regression models, RF achieves the best BP estimation performance. 
The RMSE of RF is 3.91 and 3.02 mmHg for SBP and DBP estimation, MAE is 2.17 and 1.71 mmHg, and R is 0.90 and 0.89, respectively.
\begin{table}
	\centering
	\caption{BP estimation performance of different regression models.}\label{table5}
	\begin{tabular}{cccc}
		\toprule
		Models & Evaluation metrics & SBP & DBP\\
		\midrule
		\multirow{3}*{LR} & MAE (mmHg) &8.14 $\pm$ 0.38 & 6.61 $\pm$ 0.33\\
		 \\[-3mm]\cline{2-4}	\\[-2mm]
		 & RMSE (mmHg) & 10.40 $\pm$ 0.56 & 8.28 $\pm$ 0.37  \\
		 \\[-3mm]\cline{2-4}	\\[-2mm]
		 & R & 0.41 $\pm$ 0.06 & 0.30 $\pm$ 0.03 \\
		\midrule
 		\multirow{3}*{SVR} & MAE (mmHg) & 3.11 $\pm$ 0.31  & 2.59 $\pm$ 0.26 \\
 		\\[-3mm]\cline{2-4}	\\[-2mm]
		 & RMSE (mmHg) & 5.58 $\pm$ 0.85 & 4.45 $\pm$ 0.56 \\
		 \\[-3mm]\cline{2-4}	\\[-2mm]
		 & R & 0.82 $\pm$ 0.05 & 0.79 $\pm$ 0.04 \\
		 \midrule
 		\multirow{3}*{DT} & MAE (mmHg) & 2.20 $\pm$ 0.40 & 2.02 $\pm$ 0.24 \\
 		\\[-3mm]\cline{2-4}	\\[-2mm]
		 & RMSE (mmHg) & 6.04 $\pm$ 1.14 & 5.07 $\pm$ 0.46 \\
		 \\[-3mm]\cline{2-4}	\\[-2mm]
		 & R & 0.79 $\pm$ 0.07 & 0.73 $\pm$ 0.04 \\
		 \midrule
 		\multirow{3}*{RF} & MAE (mmHg) & 2.17 $\pm$ 0.18 & 1.71 $\pm$ 0.18 \\
 		\\[-3mm]\cline{2-4}	\\[-2mm]
		 & RMSE (mmHg) & 3.91 $\pm$ 0.53 & 3.02 $\pm$ 0.46 \\
		 \\[-3mm]\cline{2-4}	\\[-2mm]
		 & R & 0.90 $\pm$ 0.02 & 0.89 $\pm$ 0.02 \\
		 
		\bottomrule
	\end{tabular}
\end{table}
\begin{table}[!t]
	\centering
	\caption{Comparison of our method with AAMI standards. }\label{table_AAMI}
	\renewcommand\arraystretch{1.5}
	\begin{tabular}{c|c|c|c|c}
		\hline
		 \multirow{2}*{} & \multicolumn{2}{c|}{SBP}&
		 \multicolumn{2}{c}{DBP}\\
		 \cline{2-5}
		   & Results & AAMI  & Results & AAMI\\
		 
	     \hline
		 ME (mmHg) & 0.08 & $\leq$ 5 & 0.01 & $\leq$ 5 \\
		 RMSE (mmHg) & 4.11 & $\leq$ 8 & 3.36 & $\leq$ 8 \\
		 
		\hline
 
	\end{tabular}
\end{table}

\begin{table}[!t]
	\centering
	\caption{Comparison of our method with BHS standards. }\label{table_BHS}
	\renewcommand\arraystretch{1.5}
	\begin{tabular}{c|c|c|c|c}
		\hline
		 \multicolumn{2}{c|}{} & \multicolumn{3}{c}{Cumulative Error Percentage (CP)}\\
		 \cline{3-5}
		 \multicolumn{2}{c|}{}  & $\leq$5mmHg & $\leq$10mmHg & $\leq$15mmHg\\
		 
			\hline
		 \multirow{2}*{Results} & SBP & 83.1\% & 95.0\% & 98.6\% \\
		 & DBP & 86.9\% & 97.7\% & 99.1\% \\
		 
		\hline
 		 \multirow{3}*{BHS standards} & Grade A & $\geq$ 60\% & $\geq$ 85.0\% & $\geq$ 95\% \\
		 & Grade B & $\geq$ 50\% & $\geq$ 75.0\% & $\geq$ 90\% \\
 		 & Grade C & $\geq$ 40\% & $\geq$ 65.0\% & $\geq$ 85\% \\
		\hline
 
	\end{tabular}
\end{table}

Table \ref{table_AAMI} and Table \ref{table_BHS} show the comparison of our methods (using RF model) with AAMI and BHS standards. 
The ME and RMSE of our proposed method are less than 5 and 8 mmHg, respectively (both the SBP and DBP satisfy), which suggests our method passes the AAMI standard. 
Further, for SBP estimation, CP ($\leq$ 5 mmHg) is 82.8\%, CP ($\leq$ 10 mmHg) is 94.3\% and CP ($\leq$ 15 mmHg) is 98.2\%, respectively. 
For DBP estimation, CP ($\leq$ 5 mmHg) is 87.9\%, CP ($\leq$ 10 mmHg) is 97.2\% and CP ($\leq$ 15 mmHg) is 99.1\%, respectively. 
Results suggest the performance of our method achieves the A level of BHS standard both for SBP and DBP estimation.

\section{Discussion}

\subsection{Comparison with Previous Studies}
In this paper, we investigate the feasibility of using brain BIOZ for BP estimation and present BrainZ-BP.
Table \ref{table7} illustrates the comparison of our proposed method with other existing studies using BIOZ for BP estimation. 
Ibrahim \emph{et al.} \cite{8863984} placed BIOZ sensor and PPG on the wrist of participants.
They extracted a total of fifty features, including PTT, time features, amplitude features, and area features.
Subsequently, they employed AdaBoost regression model for BP estimation.
Ten healthy subjects, aged between 18 and 30 years, were recruited for the experiments.
Their method achieved RMSE and MAE of 3.44 and 2.51 mmHg for SBP estimation, and achieved 2.63 and 1.95 mmHg for DBP estimation, respectively. 
Huynh \emph{et al.} \cite{8438854} extracted PTT features from wrist BIOZ and PPG signals.
The relationship between PTT and BP was determined by employing a quadratic regression model.
Fifteen healthy subjects, with an average age of 29 years, were recruited for the study.
Their reported RMSE of SBP and DBP are 8.47 $\pm$ 0.91 mmHg and 5.02 $\pm$ 0.73 mmHg, respectively. 
Huynh \emph{et al.} \cite{s18072095} also proposed to position two BIOZ sensors on the wrists of participants to measure PTT. They then utilized the PTT feature, along with the inverse relationship between PTT and PWV, to estimate BP.
Wang \emph{et al.} \cite{9247300} proposed a continuous BP monitoring system leveraging single-channel wrist BIOZ, and they also build a quadratic regression model for BP estimation.
Thirty subjects, with an average age of 27 years, participated in the experiments. 
Their reported MAE are 2.01 $\pm$ 1.40 mmHg and 2.26 $\pm$ 1.43 mmHg for SBP and DBP, respectively.
Ibrahim et al. \cite{ibrahim2022cuffless} utilized a wristband BIOZ sensor and employed a CNN autoencoder for BP estimation. 
In the experiments conducted on a sample of four subjects aged between 20 and 25 years, they achieved a RMSE of 6.5 mmHg for SBP estimation and a RMSE of 5.0 mmHg for DBP estimation.
Sel et al. \cite{sel2023continuous} developed a ring-BIOZ-based cuff-less BP estimation device, utilizing 15 BIOZ features and an AdaBoost regression model.
They recruited 10 health subjects in their mid-twenties for experiments and achieved an RMSE of 5.27 mmHg for SBP estimation and 3.87 mmHg for DBP estimation.

We can see that our proposed method achieves higher estimation performance than studies \cite{8438854,s18072095}, and achieves similar performance as studies \cite{8863984,9247300}.
The RMSE and R of the proposed method surpass the method in \cite{8438854} with an improvement of 4.41 mmHg and 0.02 for SBP estimation, and of 1.77 mmHg and 0.01 for DBP estimation, respectively.
Further, the RMSE and R of our method outperform the method in \cite{s18072095} with an improvement of 3.41 mmHg and 0.09 for SBP, and of 1.92 mmHg and 0.05 for DBP, respectively.
Since all the aforementioned studies adopted the subject-dependent paradigm to conduct experiments, the comparison in this study is fair.
Results show that brain BIOZ is a promising technique for BP estimation. 
\begin{table*}[!h]
	\centering
	\caption{Comparison of our proposed method with existing studies using BIOZ for BP estimation.
 }\label{table7}
	\begin{tabular}{cccccccccc}
		\toprule
		\multirow{2}*{Studies} & \multirow{2}*{Subjects} & \multirow{2}*{Signals} & 
  \multirow{2}*{Models} &
  \multicolumn{3}{c}{SBP}&
		\multicolumn{3}{c}{DBP}\\
		\\[-3mm]\cline{5-10}	\\[-2mm]
		& & & & MAE & RMSE & R & MAE & RMSE & R \\
		
		\midrule
		Ref. \cite{8863984} & 10 & Wrist BIOZ + PPG & AdaBoost  & 2.51 & 3.44 & 0.86 & 1.95 & 2.63 & 0.77 \\
		Ref. \cite{8438854} & 15 & Wrist BIOZ + PPG & Quadratic Regression  & - & 8.47 & 0.88 & - & 5.02 & 0.88 \\
		Ref. \cite{s18072095} & 15 & Wrist BIOZ & PWV model  & - & 7.47 & 0.81 & - & 5.17 & 0.84 \\
		Ref. \cite{9247300} & 30 & Wrist BIOZ & Quadratic Regression  & 2.01 & - & 0.95 & 2.26 & - & 0.75 \\

		Ref. \cite{ibrahim2022cuffless} & 4 & Wrist BIOZ & CNN autoencoder  & - & 6.5 & 0.79 & - & 5.0 & 0.80 \\
  
  		Ref. \cite{sel2023continuous} & 10 & Ring BIOZ & AdaBoost  & - & 5.27 & 0.76 & - & 3.87 & 0.81 \\

		This work & 13 & Brain BIOZ + ECG & RF  & 2.17 & 3.91 & 0.90 & 1.71 & 3.02 & 0.89 \\
		\bottomrule
	\end{tabular}
\end{table*}


\subsection{Effect of Excitation Frequency}
We carry out an additional experiment to investigate the influence of excitation frequency on brain BIOZ measurement.
Excitation frequency changes from 1 kHz to 20 kHz. 
The difference between the maximum and minimum of brain BIOZ in each cardiac cycle is denoted as $\Delta Z$.
$\Delta Z$ and SampEn are used as indicators to evaluate the performance of brain BIOZ measurement in this study.
Larger $\Delta Z$ can better reflect the CBF change.
SampEn can reflect the irregularity of time series.
A lower SampEn value means the higher signal quality of the measured brain BIOZ.

\begin{figure}[!t]
	\centerline{\includegraphics[scale=0.19]{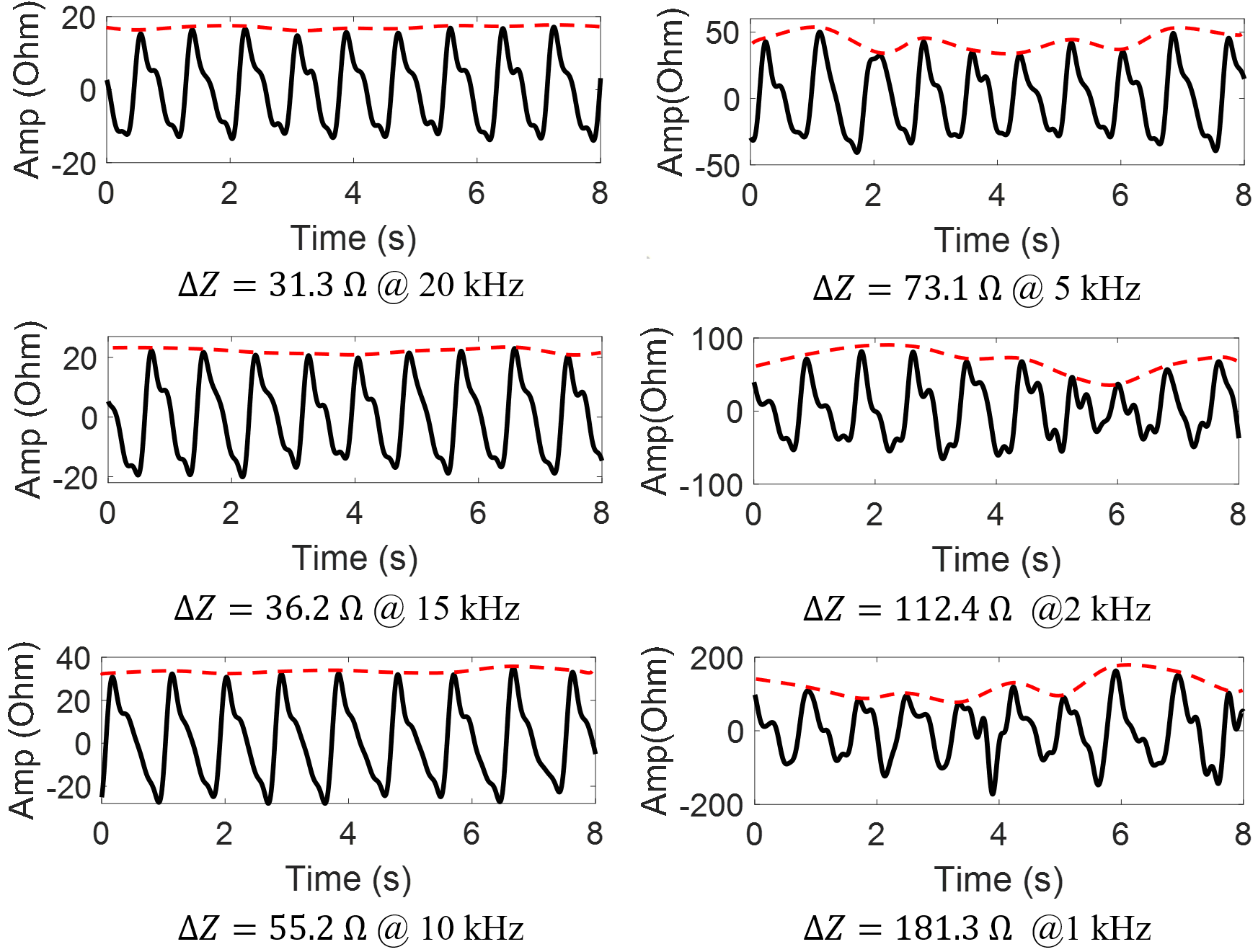}}
	\caption{Brain BIOZ waveform from different excitation frequency.}
	\label{excitation_f_wave}
\end{figure}

\begin{figure}[!t]
	\centerline{\includegraphics[scale=0.14]{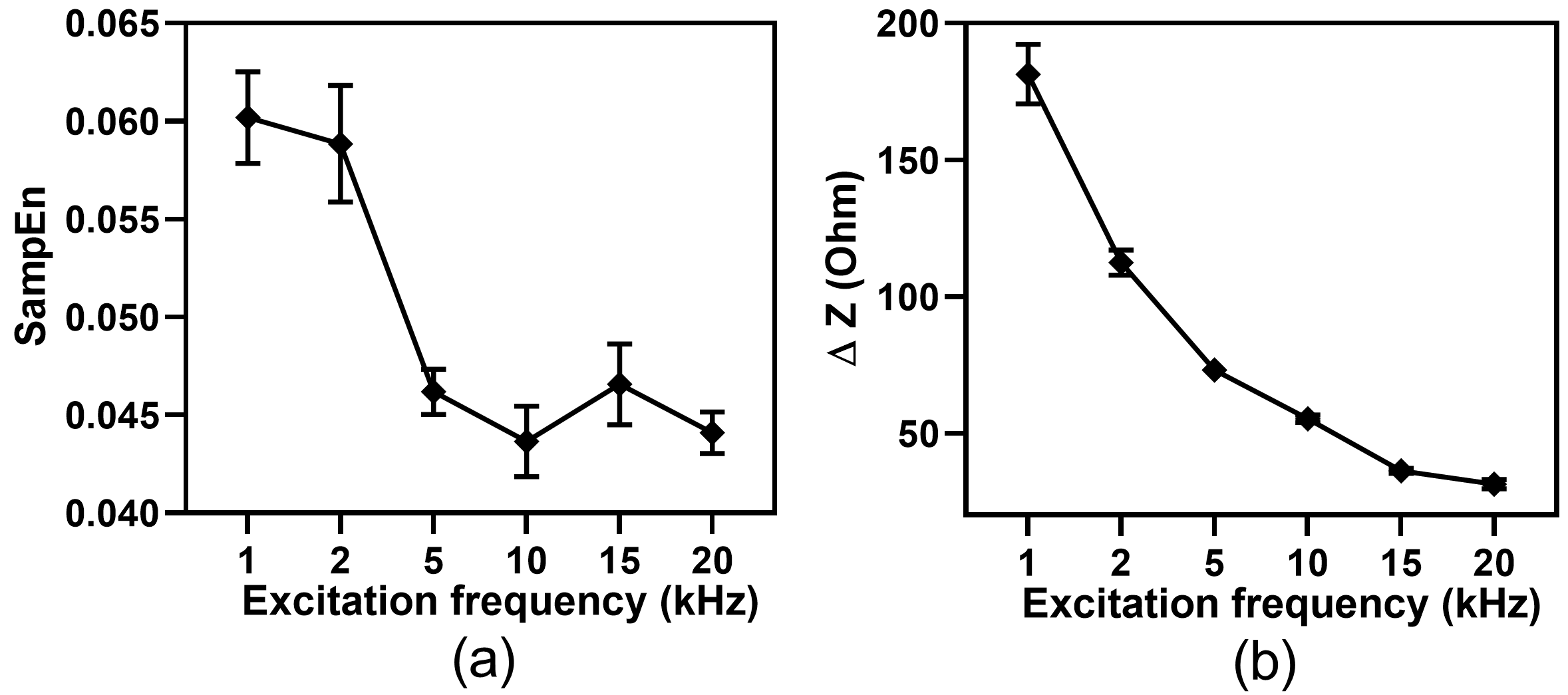}}
	\caption{SampEn and $\Delta Z$ of brain BIOZ from different excitation frequency. (a) SampEn. (b) $\Delta Z$.}
	\label{SampEn}
\end{figure}

Fig. \ref{excitation_f_wave} shows the waveform of brain BIOZ from different excitation frequencies.
It can be seen that lower excitation frequency has larger $\Delta Z$, e.g. $\Delta Z$ is 31.3 $\Omega$ and 55.2 $\Omega$ when excitation frequencies are 20 kHz and 10 kHz, respectively.
However, when excitation frequency decreases to 2 kHz, the waveform of brain BIOZ contains obvious fluctuation.
When excitation frequency decreases to 1 kHz, the waveform of BIOZ has poor regularity due to the great effect of skin-electrode impedance.
We utilize data in 100 cardiac cycles to calculate the mean and SD of SampEn and $\Delta Z$, as shown in Fig. \ref{SampEn}.
As can be seen, $\Delta Z$ increases as excitation frequency decreases.
However, when excitation frequency decreases to 2 kHz, SampEn of brain BIOZ increases largely.
The mean SampEn is 0.060 and 0.058 for 1 kHz and 2 kHz, respectively, while the mean SampEn is only about 0.045 for 5 - 20 kHz, where 10 kHz excitation frequency has the best signal quality with 0.043 mean SampEn.

Results demonstrate that lower excitation frequency has a larger $\Delta Z$.
However, too low excitation frequency will bring larger skin-electrode impedance, which may lead to low signal quality of the measured brain BIOZ.
10 kHz excitation frequency is able to produce relatively large $\Delta Z$, and is less affected by skin-electrode impedance. 
Therefore, 10 kHz excitation frequency is used in this study.

\subsection{Effect of Electrode Position}
Electrode position plays an essential role in BIOZ measurement \cite{8438854,9247300}.
Therefore, we investigate the effect of electrodes placed in the anterior-posterior direction and in the left-right direction on brain BIOZ measurement.
For anterior-posterior direction placement, two electrodes are placed on the forehead and the occipital bone of the human head, respectively.
For left-right direction placement, two electrodes are placed on the left and right temple, respectively.
As can be seen from Table \ref{BIOZ_excitation_f_tab}, 
anterior-posterior direction placement of electrodes can provide larger $\Delta Z$ than left-right direction under the excitation frequency of 2 kHz to 20 kHz.
For 10 kHz excitation frequency, anterior-posterior direction and left-right direction have almost the same SD of $\Delta Z$ (3.96 $\Omega$ for anterior-posterior, and 3.97 $\Omega$ for left-right direction), but anterior-posterior direction can obtain larger mean of $\Delta Z$ (55.27 $\Omega$ for anterior-posterior direction, and 37.35 $\Omega$ for left-right direction).
Therefore, two electrodes are placed on the forehead and occipital bone of the human head in this study.
\begin{table}
	\centering
	\caption{Mean $\pm$ SD of $\Delta Z$ using anterior-posterior and left-right direction electrodes placement.}\label{BIOZ_excitation_f_tab}
	\begin{tabular}{ccccccc}
		\toprule
	    Electrodes&\multirow{2}*{Index}&\multirow{2}*{2 kHz}&\multirow{2}*{5 kHz} &\multirow{2}*{10 kHz} & \multirow{2}*{15 kHz}& \multirow{2}*{20 kHz}\\
	    placement & & & & & &\\
		\midrule
		Anterior-  & Mean & 112.49 &73.14& 55.27 & 36.27& 31.38 \\
		\\[-3mm]\cline{2-7}	\\[-2mm]
		posterior  & SD & 10.84  & 6.14 & 3.96 & 2.51 &3.37  \\
		
		\midrule
		Left-  & Mean & 71.06 &49.38& 37.35 & 26.63& 23.76 \\
		\\[-3mm]\cline{2-7}	\\[-2mm]
		right  & SD & 7.18  & 4.95 & 3.97 & 1.81 &2.55  \\
		
		\bottomrule
	\end{tabular}
\end{table}

\subsection{Limitations and Future Works}
However, there are some limitations of this work.
In this pioneering study, we recruited 13 young subjects without any history of cardiovascular diseases to participate in the experiments.
Like many existing studies on BIOZ-based BP estimation \cite{8863984,8438854,s18072095,9247300}, we also adopted the subject-dependent paradigm for our experiments.
It means BrainZ-BP is a personalized model that requires a calibration process in practical application for each user, i.e. it requires collecting data for each user in advance for model training before the model works, no matter in the case of individuals with health conditions or advanced age.
In the future, we plan to recruit a larger number of subjects encompassing a broader range of ages and clinical conditions to perform the subject-independent experiment and enhance the generalizability of BrainZ-BP.
Secondly, in this pilot study, we only collected BP data from subjects. 
In future work, we plan to perform experiments to collect the BP and ICP data simultaneously and study a regression model that leverages brain BIOZ to estimate BP and ICP simultaneously.
Finally, recent studies have introduced the utilization of magnetic sensors, such as magnetocardiography, for the contactless monitoring of BIOZ \cite{wang2017magnetic} and cardiac signal \cite{liao2021coherent}.
BrainZ-BP monitoring system can also expand to using magnetic sensors for contactless BP estimation.
We leave this as our future work.


\section{Conclusion}

In this paper, we explore the feasibility of using brain BIOZ for BP estimation and present a novel cuff-less BP estimation model called BrainZ-BP.
BrainZ-BP utilizes brain BIOZ and ECG signals for BP estimation.
Various features including PTT-based and morphological features of brain BIOZ are extracted.
We use the Pearson correlation coefficient and random forest impurity methods to select the top 25 important features. 
The selected features are fed into the random forest regression model for BP estimation.
Results show that the MAE, RMSE, and R of BrainZ-BP are 2.17 mmHg, 3.91 mmHg, and 0.90 for SBP estimation, and are 1.71 mmHg, 3.02 mmHg, and 0.89 for DBP estimation. 
The proposed BrainZ-BP both satisfies AAMI and BHS standards. 
Results show that brain BIOZ is a promising technique for BP estimation. 
The presented BrainZ-BP model can be applied in the brain BIOZ-based non-invasive ICP monitoring scenario to monitor BP simultaneously.


\bibliographystyle{ieeetr} 
\bibliography{IEEEabrv,Main} 











\vfill

\end{document}